\documentclass[conference]{IEEEtran}
\IEEEoverridecommandlockouts
\usepackage{cite}
\usepackage{amsmath,amssymb,amsfonts}
\usepackage{algorithmic}
\usepackage{graphicx}
\usepackage{textcomp}
\usepackage{xcolor}
\def\BibTeX{{\rm B\kern-.05em{\sc i\kern-.025em b}\kern-.08em
    T\kern-.1667em\lower.7ex\hbox{E}\kern-.125emX}}
\begin{document}

\title{MaD: Mapping and debugging framework for implementing deep neural network onto a neuromorphic chip with crossbar array of synapses\\
\thanks{This research is supported by Programmatic grant no. A1687b0033 from the Singapore government’s Research, Innovation and Enterprise 2020 plan (Advanced Manufacturing and Engineering domain). A version of this paper is submitted to IJCNN 2019.}
}
%1\textsuperscript{st}
\author{\IEEEauthorblockN{ Roshan Gopalakrishnan}
\IEEEauthorblockA{\textit{Institute for Infocomm Research (I2R)} \\
\textit{ASTAR}\\
Singapore \\
roshan@i2r.a-star.edu.sg}
\and
% 2\textsuperscript{nd}
\IEEEauthorblockN{ Ashish Jith Sreejith Kumar}
\IEEEauthorblockA{\textit{School of Electrical and Electronic Engineering} \\
\textit{Nanyang Technological University}\\
Singapore \\
ashishji001@e.ntu.edu.sg}
\and
% 3\textsuperscript{rd}
\IEEEauthorblockN{Yansong Chua}
\IEEEauthorblockA{\textit{Institute for Infocomm Research (I2R)} \\
\textit{ASTAR}\\
Singapore \\
chuays@i2r.a-star.edu.sg}
%\and
%\and
%%\and
}

\maketitle

\begin{abstract}
Neuromorphic systems or dedicated hardware for neuromorphic computing is getting popular with the advancement in research on different device materials for synapses, especially in crossbar architecture and also algorithms specific or compatible to neuromorphic hardware. Hence, an automated mapping of any deep neural network onto the neuromorphic chip with crossbar array of synapses and an efficient debugging framework is very essential. Here, mapping is defined as the deployment of a section of deep neural network layer onto a neuromorphic core and the generation of connection lists among population of neurons to specify the connectivity between various neuromorphic cores on the neuromorphic chip. Debugging is the verification of computations performed on the neuromorphic chip during inferencing. Together the framework becomes Mapping and Debugging (MaD) framework. MaD framework is quite general in usage as it is a Python wrapper which can be integrated with almost every simulator tools for neuromorphic chips. This paper illustrates the MaD framework in detail, considering some optimizations while mapping onto a single neuromorphic core. A classification task on MNIST and CIFAR-10 datasets are considered for test case implementation of MaD framework.    
\end{abstract}


\section{Introduction}

Edge computing is one of the recent developments in the field of artificial intelligence. The amount of data being processed with the ever increasing inter-connectivity of devices and internet of things, ranging from sensors to autonomous vehicles, demand for high real time data processing at the edge. The edge devices are usually selected from neuromorphic chips, embedded devices, FPGA, GPU/CPU etc depending on the application. Among these devices, neuromorphic chip has proven to be the efficient or potential candidate in terms of computational power and latency. Neuromorphic chips are developed in digital \cite{spinnaker}, analog or mixed signal \cite{neurogrid} \cite{brainscales} \cite{truenorth} integrated circuit designs. Usual design trend is that mostly the computation and memory section is done in analog domain whereas, the communication between cores are maintained in digital domain.

The neuromorphic chip discussed in this paper is based on crossbar architecture \cite{crossbar} of non volatile memory synapses. However, one of the main challenges is to efficiently map the neurons on to the neuromorphic chip with hardware constraints such as core size, number of cores and fan-in/fan-out \cite{neutrams}. The existing neuromorphic chips have a mapping framework which is more hardware specific. IBM's TrueNorth chip \cite{truenorth_TCAD} uses corelet language \cite{IBM_corelet} based on MATLAB, a programming language specific to their hardware. Within this MATLAB framework, a mapping technique is integrated as a minimization problem \cite{truenorth_TCAD}. SpiNNaker and BrainScaleS uses a simulator-independent language, PyNN \cite{PyNN} based on Python. Sequential mapping is used in SpiNNaker. Neural engineering framework (NEF) is developed for Neurogrid \cite{NEF}. Neutrams \cite{neutrams} addresses an optimized mapping technique based on graph partition problem: Kernighan-Lin (KL) partitioning strategy for network on chips. Even though, every neuromorphic chip simulator tools are addressing certain mapping techniques, optimized mapping onto a single neuromorphic core is often neglected and left unexplored by default. Most of these mapping techniques are hidden within a neuromorphic hardware specific simulators, which mitigate the requirement of an algorithm developer to understand the details of a neuromorphic chip. But, for an optimized co-development of a neural network model for a specific neuromorphic chip, the knowledge of hardware constraints is a must.

Over the years, convolutional neural networks evolved to become more deep and wide with respect to the evolution of different classication tasks i.e. from simple MNIST handwritten digit classification to much more complex ImageNet image classification. For MNIST classification task, as the neural network is small, the neurons can be mapped manually onto a neuromorphic core. But, for large networks in the case of ImageNet classification, it is near impossible to manually mention how the neurons in every layers are mapped to each core in a neuromorphic chip. Hence, an automated procedure is necessary for identifying the neuron addresses with corresponding synaptic weights and input values.

In this paper, aforementioned issues are mitigated with the help of MaD framework and its optimizations. MaD framework is a generic Python wrapper which has an optimized algorithm for mapping any feed forward neural network such as MLP, CNN, SNN onto a crossbar array of synapses with corresponding synaptic weights, thereby fitting the neurons in minimum possible number of neuromorphic cores. Python wrapper is also suitable as a debugging tool for verification of the inferencing of neural network architectures on the neuromorphic chip. Thus together the framework is called as mapping and debugging (MaD) framework. This Python wrapper is developed in connection with the simulator in \cite{neurosim}, where most of the techniques are quite similar to Neutrams \cite{neutrams}.

The paper is organized as follows. Section \ref{MaM} briefly describe about the crossbar array of synapses and the spiking neuron in a neuromorphic chip. Section \ref{EF} illustrates the details of MaD framework. Section \ref{Res} shows the implementation of MNIST and CIFAR-10 classification task on MaD framework. Finally the paper is concluded with discussion in section \ref{DaC}. 

\section{Materials and Method}
\label{MaM}

\subsection{Spiking Neuron}
\label{Spike_Neuron}

\begin{figure}[htbp]
\centerline{\includegraphics[scale=0.5]{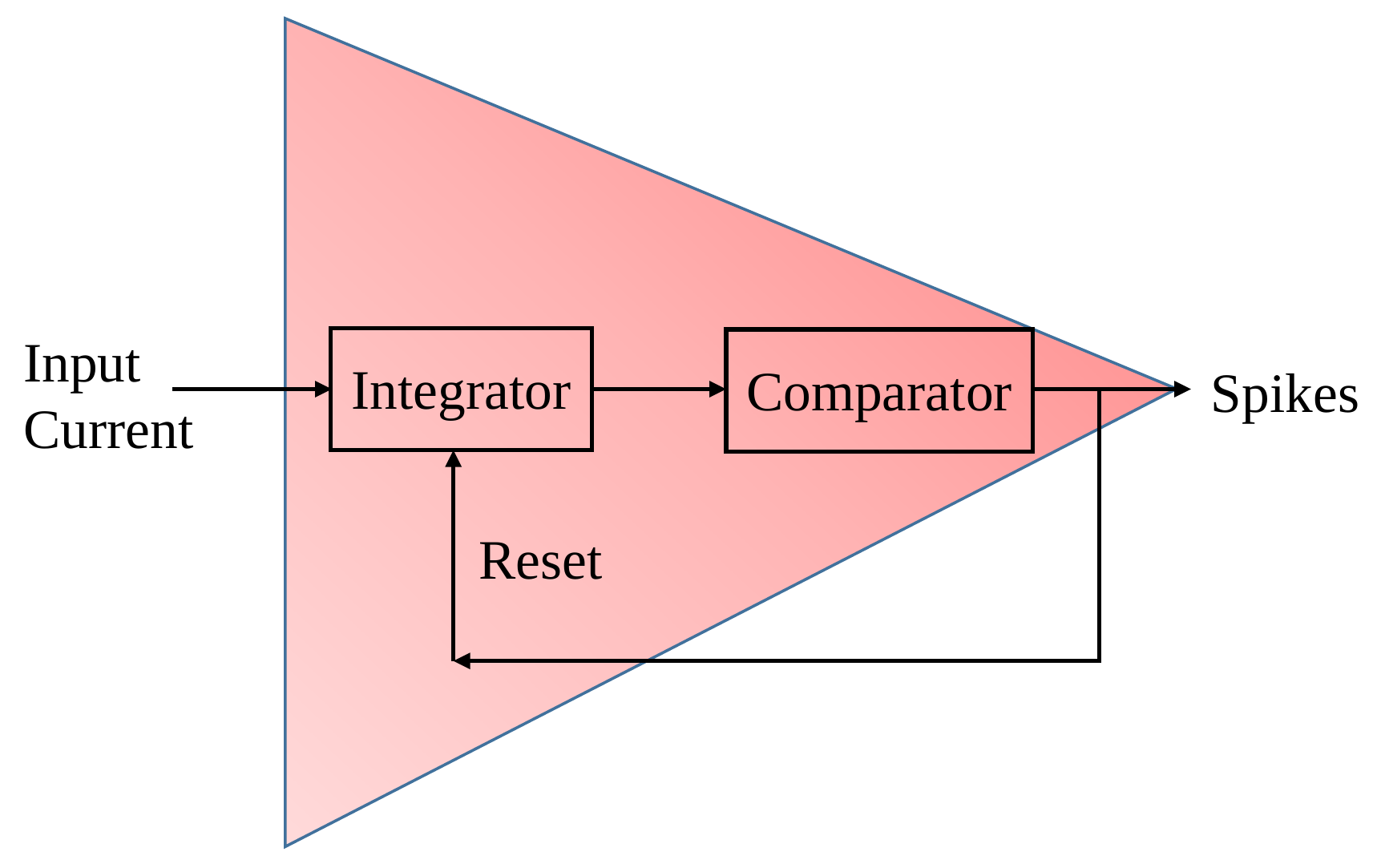}}
\caption{Block diagram of a spiking neuron.}
\label{Spi_neu}
\end{figure}

Biological neuron computes the signal received through multiple dendrites and transmits the output signal through axons to other neurons connected in the network \cite{ANS}. Fig \ref{Spi_neu} shows a block diagram representation of the biological neuron. Neuron has mainly two blocks, an integrator and a comparator. The integrator sums up all the input currents (excitatory post-synaptic current, EPSC) and build up the membrane potential. This membrane potential is being monitored by the comparator to cross certain threshold. If the membrane potential crosses the set threshold, neuron emits an output spike and then resets the membrane potential back to its initial value. The communication between neurons in the biological network or in a spiking neural network (SNN) is with the help of these output spikes. The entire mechanism of a spiking neuron can be modelled with the leaky integrate and fire neuron model and its mathematical expression \cite{Neu_dyn} is given below:

\begin{equation}
\begin{aligned}
	\label{eq:neuron}
	\tau_m \frac{du}{dt} = -[u(t)-u_{rest}] + RI(t)
\end{aligned}
\end{equation}
Where, \\
$\tau_m$ = RC, is the membrane time constant of leaky integrator.\\
u(t) = membrane potential \\
I(t) = synaptic current \\
$u_{rest}$ = membrane resting potential \\

%\subsection{RRAM Synapse}
%\label{RRAM_syn}

%\begin{figure}[htbp]
%\centerline{\includegraphics[scale=0.58]{/home/roshan/Documents/My_Publications/IJCNN_2019/figures/RRAM_synapse.pdf}}
%\caption{RRAM synapse between an axon and a neuron.}
%\label{RRAM_synapse}
%\end{figure}

%RRAM is a two terminal non volatile device with a conducting dielectric layer sandwiched between two electrodes as shown in fig. \ref{Crossbar}. Manipulation of oxygen vacancies in the conductance layer using positive and negative voltages helps in controlling current flow in RRAM. The state of RRAM reflects the current passed through it in the history, making it useful for modelling the synaptic weights of neurological synapses and implementing neural network architectures. In the case of spiking neural network (SNN), the tunable resistive state of RRAM synapses is analogous to the synaptic plasticity in brain. The electrical connection between a presynaptic neuron and a postsynaptic neuron (as shown in fig) changes, strengthening or weakening the synaptic impulses thus making it a case for brain-like pattern recognition.

\subsection{Crossbar Array of Synapses}
\label{C_RRAM}

\begin{figure}[htbp]
\centerline{\includegraphics[scale=0.4]{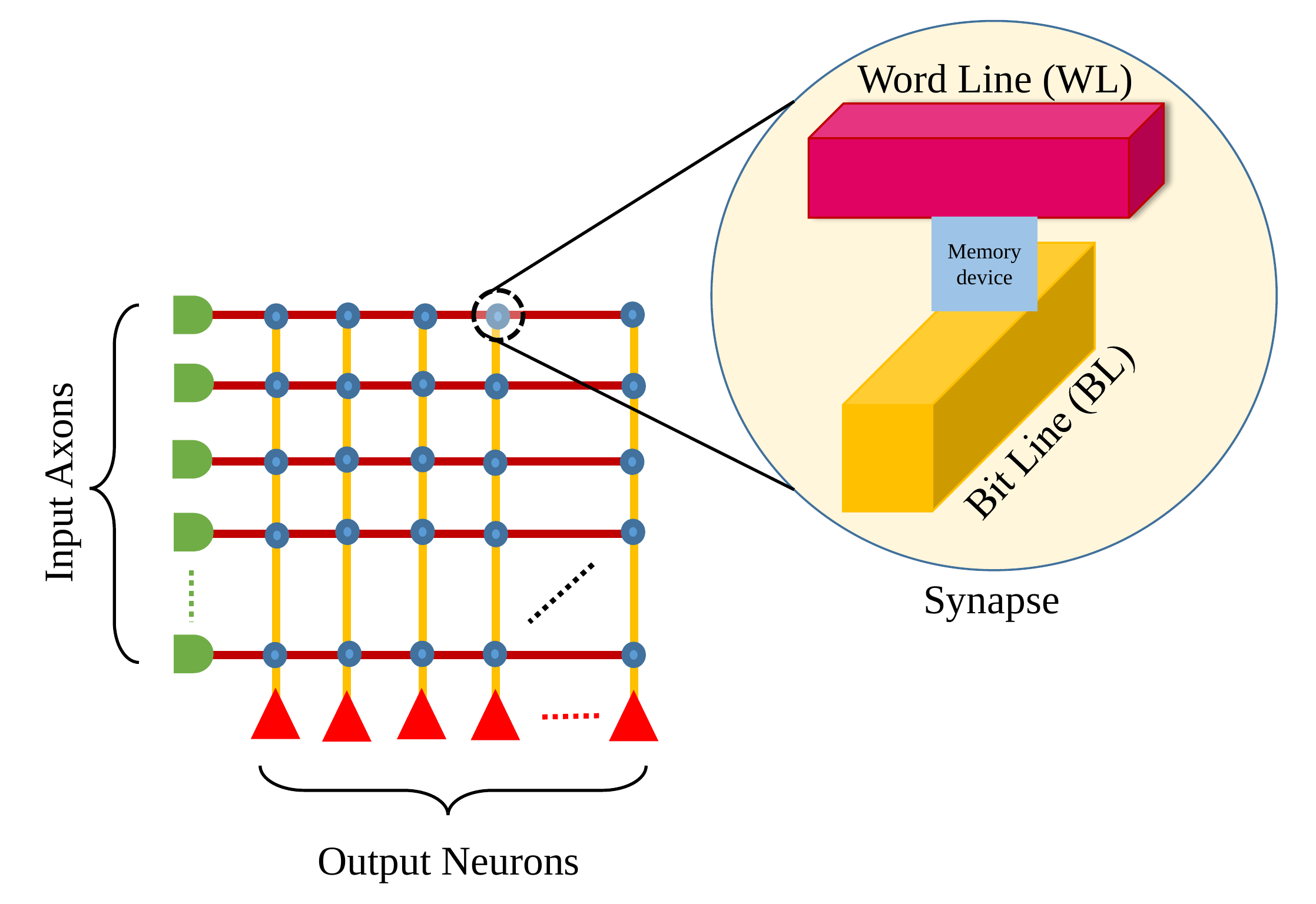}}
\caption{Crossbar array of synapses in a neuromorphic core.}
\label{Crossbar}
\end{figure}

%Crossbar array of RRAM synapses have produced promising results in neuromorphic hardware, with better on/off ratio , high retention rates , lower leakage current and fast switching between on-off states. 
Fig. \ref{Crossbar} shows a crossbar array of synapses. The crossbar structure is very suitable for performing matrix dot vector multiplication (MVM) \cite{MVM} along each column in a crossbar architecture. For instance,  a neuromorphic core with a core size of 256$\times$256, input voltages from respective axons out of 256 are given through word line. Bit line collects all the weighted current at each synaptic nodes (256$\times$256) and delivers to respective output neurons (256) for integration. The weighted current depends on the memory element used in the intersection of word line and bit line as synapse. The synaptic weights, which draws analogy to conductances, are represented in the form of blue dots at the cross points. From Kirchoff’s current law, the total current flowing into each neuron from respective bit lines is the sum of currents flowing through each intersection in every column. In fact, in  conventional neural networks, total current of a particular column is the value of a single neuron activation in a particular layer, formed by summation of products of input voltages and corresponding synaptic weights (conductances) taking part in convolution operation. %The RRAM device is capable of being stacked into different layers making 3D architectures possible which ensures high computational parallelism.

\section{MaD Framework}
\label{EF}

\begin{figure}[htbp]
\centerline{\includegraphics[width=9cm,height=6cm]{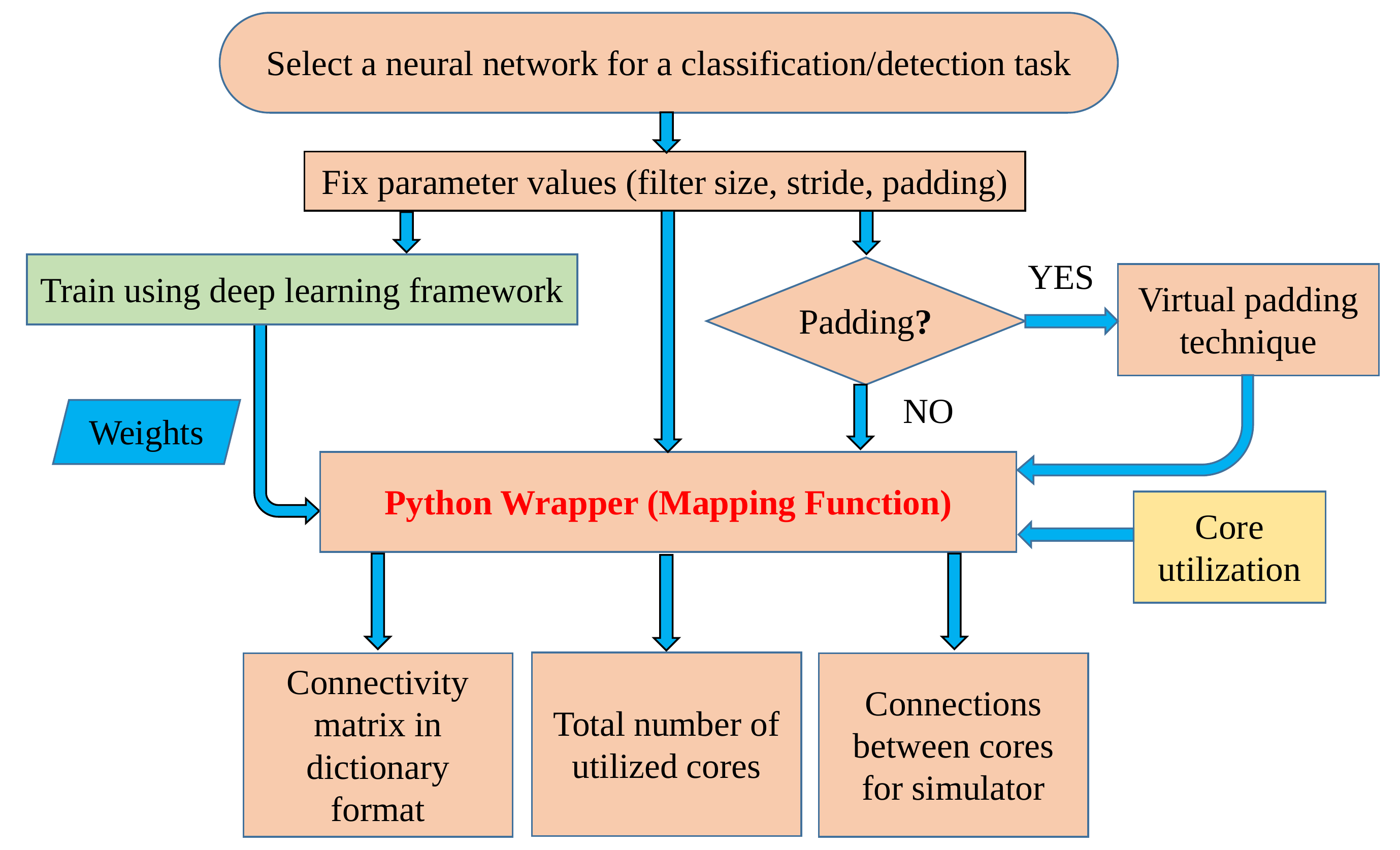}}
\caption{\textbf{Flowchart of Python wrapper:} The details of the Python wrapper is shown with a flowchart. The input and output of the mapping function that is used in the python wrapper is illustrated in the flowchart. The core utilization and weight files are marked in different color to show that these inputs are the results from other functions.}
\label{Flowchart}
\end{figure}

This section illustrates the details of the construction of MaD framework. The complete usage of the framework is explained with a flowchart as shown in fig. \ref{Flowchart}. A particular neural network is chosen for a classification or a detection task. The parameters like filter size, strides and padding among each layers are fixed. The chosen network is trained using deep learning tool for obtaining the weight files to be given as input to the mapping function. Core utilization is defined as the number of axons and neurons utilized in a single neuromorphic core. Core utilization, as shown in the flowchart, is an output from another function which calculates the number of axons and number of neurons used for mapping a section of particular layer onto a single core. Core utilization is represented as [axons $\times$ neurons]. The details of the mapping function, core utilization and padding techniques are given in the subsequent subsections. This section is ended by including optimizations to be considered while mapping.

%\subsection{Flowchart}
%steps for creating MaD 

\subsection{Mapping Function}
\label{MF}

The mapping function is the core of the Python wrapper as shown in fig. \ref{Flowchart}. Fig. \ref{Flowchart} shows the input and output of the mapping function. The inputs to mapping function are input size, filter size, stride, padding, core utilization and weight files. The input size is the size of the input datasets, for eg. 28$\times$28 in the case of MNIST or 32$\times$32 in the case of CIFAR-10. Filter size is the size of filters used for convolution in each layers, here it is selected as 3$\times$3 throughout the layers of the chosen neural networks in section \ref{Res}. Stride and padding depends on the layers of the convolutional neural network. The detailed calculation of the core utilization is mentioned in subsection \ref{CU}. Weight files are the weights obtained after training the chosen neural network using deep learning tool. The output section in fig. \ref{Flowchart} shows the necessary outputs that is obtained from the mapping function. There are mainly three outputs, a connectivity matrix for verifying the interconnectivity between the cores and within the core, to verify the cores utilized and an automated generation of connection list for simulator.

The steps for mapping are as follows:
\begin{list}{•}{•}
\item 1. All the neurons are first named to follow a regular pattern eg. L1-F1-N[1,1] this implies layer:1, feature map:1, and neuron in row:1 and column:1.
\item 2. Prepare a connectivity list of population of neurons in a particular layer connected to the previous layer.
\item 3. Choose a population of neurons from a particular layer, based on the core utilization, to be mapped on to a particular core.
\item 4. Repeat this process until entire neurons in every layers are completely mapped onto the core. Since the naming and connectivity list are fixed at the beginning, the neurons and axons will be automatically duplicated among the cores for mapping.  
\end{list}
     
\subsection{Core Utilization}
\label{CU}

\begin{figure}[htbp]
\centerline{\includegraphics[width=9cm,height=6cm]{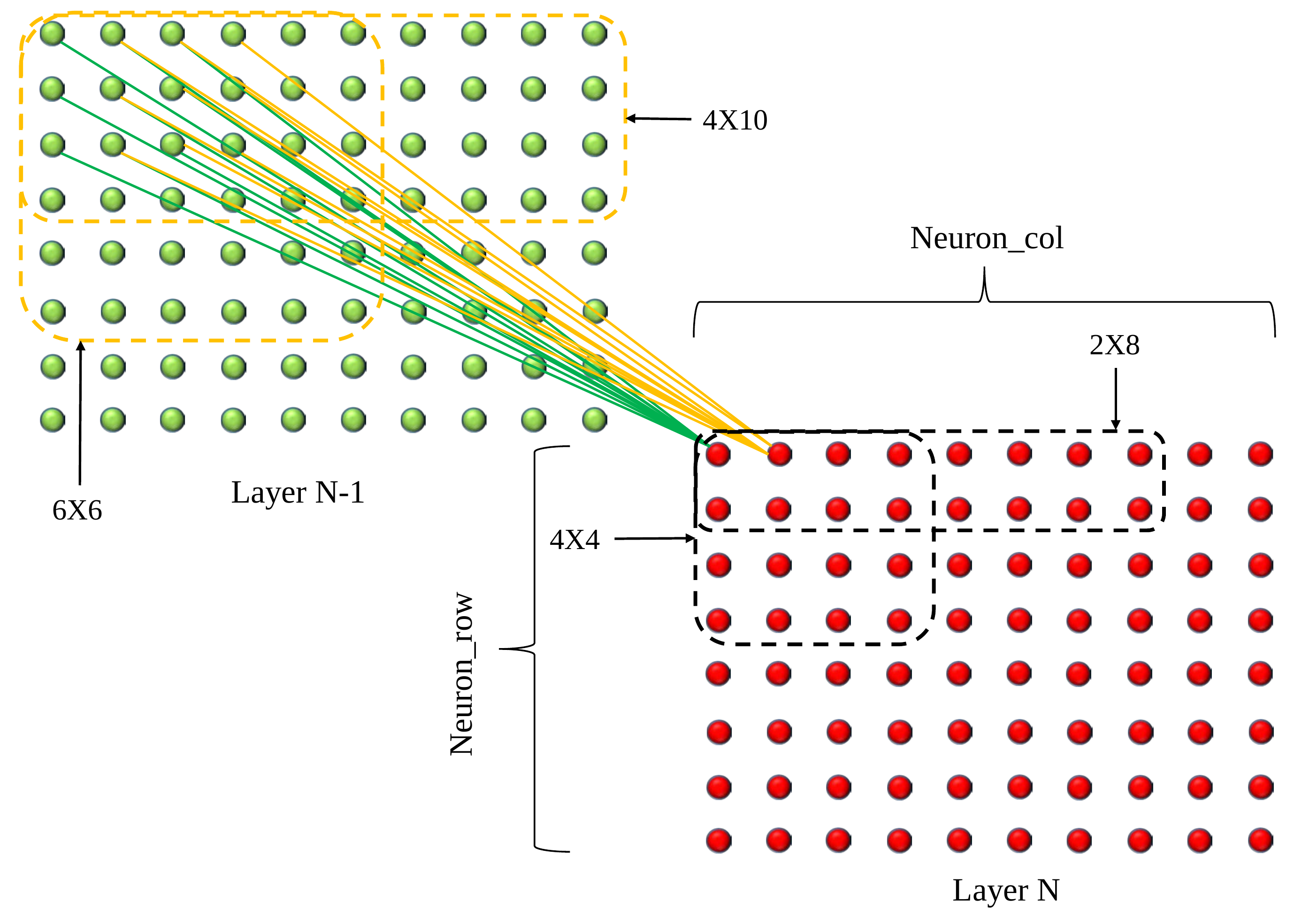}}
\caption{Two layers of convolution layer to illustrate the optimization of core utilization. Layer N-1 neurons are in green, whereas layer N neurons are in red. Synaptic connections are shown for two neurons in layer N.}
\label{Core_uti}
\end{figure}

Consider two layers of a convolutional neural network shown in fig \ref{Core_uti}. The neurons in layer N is marked as red and neurons in layer N-1 is marked as green. First two neurons in layer N is connected to layer N-1 and the synaptic connections are shown with straight lines. The convolution filter size used is 3$\times$3, hence you can see 9 connections from each red neurons in layer N to 9 green neurons in layer N-1. Likewise, the synaptic connections can be imagined throughout the layer with respect to the kernel size and strides used for convolution.  While mapping these two layers in fig. \ref{Core_uti} onto a core with crossbar array, the green neurons in layer N-1 will be the axons and the red neurons in layer N will be the neurons as in fig \ref{Crossbar}. Notice the overlap of filter window when it strides across the layer. In fact these overlapped green neurons can be mapped onto the crossbar array connections without any duplication. Duplicating the axons, while one to one mapping of neurons connected to axons onto a core, is not a good design with respect to core utilization as input needs to be duplicated into many axons and also mapping requires bigger core sizes and ends up utilizing many cores \cite{mems_IJCNN_2016}. Hence, the toeplitz matrix method is utilized for efficient mapping of these layers onto a neuromorphic core without input duplication. Toeplitz method for convolution is illustrated in \cite{toeplitz_IBM} \cite{toeplitz}. Inorder to calculate the core utilization, the number of neurons and axons connected together has to be chosen which could be entirely mapped onto a single core. The number of axons can be evaluated as an algorithmic condition in the mapping function as there are overlapping axons whereas neurons selection become bit straight forward. The overlapping axons are defined as the axons which share connections with more than a single neuron, the term overlapping is because of the overlapping nature of the axons with the neighbourhood of the kernel filter with respect to strides (see layer N-1 in fig. \ref{Core_uti}, the overlapping axons among the green and yellow synaptic connections are 3). Depending on this overlap, kernel filter size and strides, the total number of axons to be selected follows the formula as given below:

\begin{equation}
\begin{aligned}
	\label{eq:Axons}
	N\_axons = KXK + KXSX(Neuron\_col-1) + \\
		   SXSX(Neuron\_col-1)X(Neuron\_row-1) + \\
		   KXSX(Neuron\_row-1)
\end{aligned}
\end{equation}
Where, \\
$N\_axons$ = total number of axons to be selected \\
K = convolution filter size \\
S = stride \\
$Neuron\_row$ = number of neurons across row \\
$Neuron\_col$ = number of neurons across column \\

The selection of neurons, $Neuron\_row$ and $Neuron\_col$, in a layer depends on the condition: number of axons, $N\_axons$ $<=$ number of physical axons (eg. 256 or 512 or 1024) in the neuromorphic core. Eq. \ref{eq:Axons} is considering only a single feature map, this can be easily extended to multiple feature maps by multiplying with respective channel size.  

\subsection{MaD Framework Optimizations}

\subsubsection{Core Utilization}

Referring to fig. \ref{Core_uti}, consider a case for calculating core utilization, suppose 16 neurons has to be chosen from layer N for mapping onto a core. This can be done by choosing 2 rows and 8 columns of neurons or 4 rows and 4 columns of neurons. Here, rows and columns of neurons correspond to $Neuron\_row$ and $Neuron\_col$ in eq.\ref{eq:Axons}. If the convolution kernel size, K used is $3\times3$ and stride, S is 1, then for 2 rows and 8 columns of neurons the axons required are 4 rows and 10 columns, similarly for 4 rows and 4 columns of neurons the axons required are 6 rows and 6 columns. This can be easily estimated from the formula to calculate the output size of convolutions as given below:

\begin{equation}
\begin{aligned}
\label{eq:Conv}
& O\_{width} = \frac{I\_width - F\_width}{Stride\_width} + 1 \\
& O\_height = \frac{I\_height - F\_height}{Stride\_height} + 1 \\
\end{aligned}
\end{equation}
Where, \\
$O\_width \hspace{5pt} and \hspace{5pt} O\_height$ = Width and height of the convolution output respectively  \\
$I\_width \hspace{5pt} and \hspace{5pt} I\_height$ = Input width and height respectively \\
$F\_width \hspace{5pt} and \hspace{5pt} F\_height$ = Width and height of filter kernel \\
$S\_width \hspace{5pt} and \hspace{5pt} S\_height$ = Width and height of strides \\

The above case suggest that choosing neurons from 4 rows and 4 columns are much better for core utilization than from 2 rows and 8 columns as input number of axons in former case is only 36 whereas, in the later case it is 40. That means the core utilization is [36$\times$16] in the former case and [40$\times$16] in the later case. The intuition from this example case is that the neurons to be selected for mapping onto the core is better to be in square shape than in rectangular shape. The section below provides a mathematical proof for choosing square shape rather than rectangular shape while mapping:

\begin{figure}[htbp]
\centerline{\includegraphics[width=7cm,height=6cm]{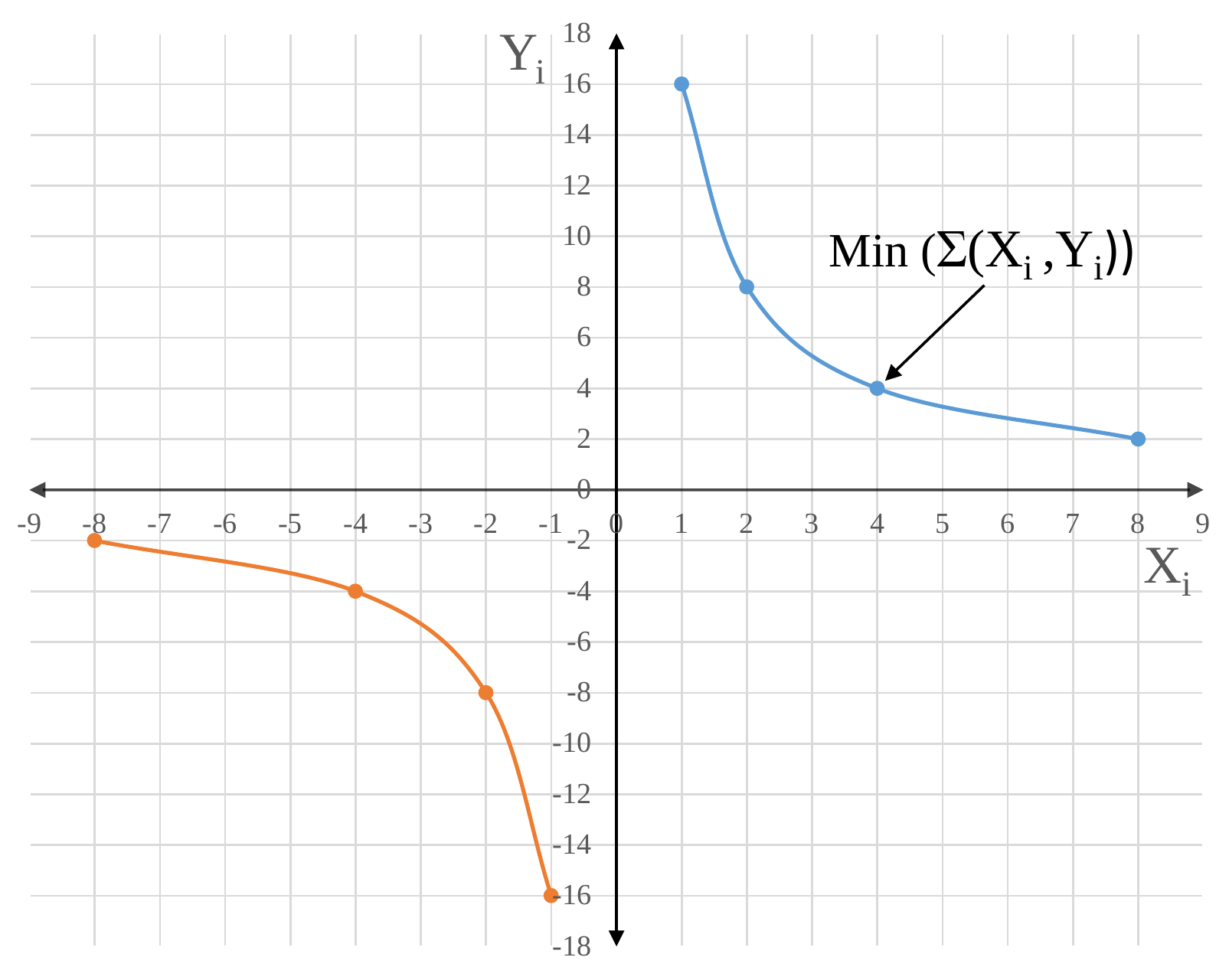}}
\caption{Graphical illustration of the theorem.}
\label{Theorem}
\end{figure}

\textit{Theorem}: Given `a', find $X_i$ and $Y_i$ such that: $X_i \times Y_i$ = a and minimum of $\sum(X_i , Y_i)$.

\textit{Proof}: The graphical illustration of the theorem is shown in fig. \ref{Theorem}. \\
%Given Z = XY \\
%If Y = 1, then X = Z and X+Y = Z+1 \\
%If Y = \sqrt{Z}, then X = \sqrt{Z} and X+Y = 2\sqrt{Z} \\
\begin{equation}
\begin{aligned}
\label{eq:theo}
& Consider \hspace{5pt} XY = a \hspace{5pt} and \hspace{5pt} X+Y = Z \\
& where \hspace{5pt} Z \hspace{5pt} is \hspace{5pt} a \hspace{5pt} real \hspace{5pt} number \\
& X + \frac{a}{X} = Z \\
& \frac{dZ}{dX} = 1 - \frac{a}{X^2} \\
& at \hspace{5pt} minima \hspace{5pt} \frac{dZ}{dX} = 0, \\
& \therefore \hspace{5pt} 1 - \frac{a}{X^2} = 0 \\
& X = +-\sqrt{a} \\
& If \hspace{5pt} X > 0, \hspace{5pt} Z \hspace{5pt} is \hspace{5pt} minimum \hspace{5pt} \therefore X = \sqrt{a} \\
& also \hspace{5pt} Y = \sqrt{a} \hspace{5pt} and \hspace{5pt} X+Y \hspace{5pt} is \hspace{5pt} the \hspace{5pt} minimum. \\
\end{aligned}
\end{equation}

\vspace{2mm}
\subsubsection{Padding}

Padding is a common technique used in deep learning for maintaining the shape of the convolution layers throughout the network. Padding simply adds extra zeros around the input activations in a convolution layer during convolution operation. In fact such added zeros doesn't provide any computational significance as mathematically zeros are multiplied and added. While mapping, these padded zeros are in fact physical neurons, but need not be participating in computation. If these neurons are considered as physical neurons during mapping, then there will be a lot of wastage on axon usage. This will reduce the optimized utilization of core. Hence, as shown in fig. \ref{Flowchart} (mentioned as virtual padding technique), when padding is used in a particular convolution layer, a virtually padded neuron address is created and is assigned in the connectivity list. Later, while mapping onto the core these virtually padded neurons are removed from the connectivity list, reducing the fan in connection of those particular neurons in the periphery of a layer connected to those padded neurons in the previous layer.

\section{Results}
\label{Res}

This section mainly provides an instance of the utilization of neuromorphic chip using a classification task on MNIST and CIFAR-10 datasets through the parameters, core utilization and number of cores utilized. Here, the focus is not on improving the accuracies, but to show the neuromorphic core utilization while mapping, with a much simpler handcrafted neural network. All the accuracies mentioned in this section is iterated for ten times and then averaged it out. Two sets of experiments are done for that purpose, one is to choose a particular neural network architecture for classification task on MNIST and CIFAR-10 datasets and keep that architecture constant among different core sizes. Different core sizes chosen here are [256$\times$256], [512$\times$512] and [1024$\times$1024] (core sizes need not be in square shape but any other shapes are also possible). Here, the accuracy will be same, as architecture is constant, while the core utilization and number of cores utilized will be different among different core sizes. Second set of experiment is to change the neural network architecture for different core sizes. This will change the accuracy of neural network architectures for different core sizes, but the number of cores utilized will remain same.

%\begin{table}[htbp]
%\caption{Neural network (NN) architecture for MNIST dataset}
%\begin{center}
%\begin{tabular}{|c|c|c|c|}
%\hline
%\textbf{NN}&\multicolumn{3}{|c|}{\textbf{Core Size}} \\
%\cline{2-4} 
% & \textbf{\textit{256$\times$256}}&  \textbf{\textit{512$\times$512}}& \textbf{\textit{1024$\times$1024}} \\
%\hline
%Layers & More table copy$^{\mathrm{a}}$& &  \\
%\textbf{Layers} & 28$\times$28$\times$1--28$\times$28$\times$8 & 28$\times$28$\times$1--28$\times$28$\times$16 & 28$\times$28$\times$1--28$\times$28$\times$32 \\
%\textbf{1-2} & K = 3$\times$3, S=1, P=1 & K = 3$\times$3, S=1, P=1 & K = 3$\times$3, S=1, P=1 \\ 
%\cline{2-4} 
%\hline
%\textbf{Layers} & 28$\times$28$\times$8--28$\times$28$\times$16 & 28$\times$28$\times$16--28$\times$28$\times$32 & 28$\times$28$\times$32--28$\times$28$\times$64 \\
%\textbf{2-3} & K = 3$\times$3, S=2, P=1 & K = 3$\times$3, S=2, P=1 & K = 3$\times$3, S=2, P=1 \\ 
%\cline{2-4} 
%\hline
%\textbf{Layers} & 28$\times$28$\times$16--28$\times$28$\times$64 & 28$\times$28$\times$32--28$\times$28$\times$128 & 28$\times$28$\times$64--28$\times$28$\times$256 \\
%\textbf{3-4} & K = 3$\times$3, S=2, P=0 & K = 3$\times$3, S=2, P=0 & K = 3$\times$3, S=2, P=0 \\ 
%\cline{2-4} 
%\hline
%\multicolumn{4}{l}{$^{\mathrm{a}}$Sample of a Table footnote.}
%\end{tabular}
%\label{tab1}
%\end{center}
%\end{table}

\begin{table}[htbp]
\caption{Neural network (NN) architecture for MNIST dataset}
\begin{center}
\begin{tabular}{|c|c|c|c|}
\hline
\textbf{NN}&\multicolumn{3}{|c|}{\textbf{Core Size}} \\
\cline{2-4} 
 \textbf{Architecture} & \textbf{\textit{256$\times$256}}&  \textbf{\textit{512$\times$512}}& \textbf{\textit{1024$\times$1024}} \\
\hline
%Layers & More table copy$^{\mathrm{a}}$& &  \\
\textbf{Input} & 28$\times$28$\times$1 & 28$\times$28$\times$1 & 28$\times$28$\times$1 \\
\cline{2-4} 
\hline
\textbf{Layer 1} & 28$\times$28$\times$8 & 28$\times$28$\times$16 & 28$\times$28$\times$32 \\ 
\cline{2-4} 
\hline
\textbf{Layer 2} & 14$\times$14$\times$16 & 14$\times$14$\times$32 & 14$\times$14$\times$64 \\
\cline{2-4} 
\hline
\textbf{Layer 3} & 6$\times$6$\times$64 & 6$\times$6$\times$128 & 6$\times$6$\times$256 \\
\cline{2-4} 
\hline
%\multicolumn{4}{l}{$^{\mathrm{a}}$Sample of a Table footnote.}
\end{tabular}
\label{tab1}
\end{center}
\end{table}

\begin{table}[htbp]
\caption{Neural network (NN) architecture for CIFAR-10 dataset}
\begin{center}
\begin{tabular}{|c|c|c|c|}
\hline
\textbf{NN}&\multicolumn{3}{|c|}{\textbf{Core Size}} \\
\cline{2-4} 
 \textbf{Architecture} & \textbf{\textit{256$\times$256}}&  \textbf{\textit{512$\times$512}}& \textbf{\textit{1024$\times$1024}} \\
\hline
%Layers & More table copy$^{\mathrm{a}}$& &  \\
\textbf{Input} & 32$\times$32$\times$3 & 32$\times$32$\times$3 & 32$\times$32$\times$3 \\
\cline{2-4} 
\hline
\textbf{Layer 1} & 30$\times$30$\times$8 & 30$\times$30$\times$16 & 30$\times$30$\times$32 \\ 
\cline{2-4} 
\hline
\textbf{Layer 2} & 14$\times$14$\times$16 & 14$\times$14$\times$32 & 14$\times$14$\times$64 \\
\cline{2-4} 
\hline
\textbf{Layer 3} & 6$\times$6$\times$64 & 6$\times$6$\times$128 & 6$\times$6$\times$256 \\
\cline{2-4} 
\hline
%\multicolumn{4}{l}{$^{\mathrm{a}}$Sample of a Table footnote.}
\end{tabular}
\label{tab2}
\end{center}
\end{table}

\subsubsection{Keeping architecture constant}

Consider the architecture shown in table \ref{tab1} and \ref{tab2}. The softmax classifier output layer is not shown in the tables. For this set of experiment, the architecture is maintained same irrespective of different core sizes -- neural network architecture chosen for MNIST and CIFAR-10 datasets are given in the first column under the core size, 256$\times$256 respectively in both tables \ref{tab1} and \ref{tab2}. The neural network architecture is kept constant while mapping onto other core sizes as well. The convolutional filter size used is 3$\times$3 throughout the layers. In table \ref{tab1}, between input layer and layer 1, stride used is 1 and with padding in the input activations. Between layer 1 and layer 2, stride used is 2 and with padding in the input. Between layer 2 and layer 3, stride used is again 2 but without padding in the input. In table \ref{tab2}, between input layer and layer 1, stride used is 1 and without any padding in the input. Between layer 1 and layer 2, stride used is 2 and without padding in the input. Between layer 2 and layer 3, stride used is again 2 but with padding in the input. From table \ref{tab3} and \ref{tab4}, the results for MNIST and CIFAR-10 classification accuracy is constant among all the core sizes as the architecture remains same, while the core utilization and number of cores utilized changes with core sizes.

\begin{table}[htbp]
\caption{Keeping architecture constant: MNIST dataset}
\begin{center}
\begin{tabular}{|c|c|c|c|c|c|c|}
\hline
\multicolumn{6}{|c|}{\textbf{Core Size}}&\textbf{Acc} \\
\cline{1-6} 
\multicolumn{2}{|c|}{\textbf{\textit{256$\times$256}}}&\multicolumn{2}{|c|}{\textbf{\textit{512$\times$512}}}&\multicolumn{2}{|c|}{\textbf{\textit{1024$\times$1024}}}&\textbf{(\%)} \\
\cline{1-6}
\textbf{\textit{Core}}&\textbf{\textit{No of}}&\textbf{\textit{Core}}&\textbf{\textit{No of}}&\textbf{\textit{Core}}&\textbf{\textit{No of}}& \\
\textbf{\textit{utilization}}&\textbf{\textit{cores}}&\textbf{\textit{utilization}}&\textbf{\textit{cores}}&\textbf{\textit{utilization}}&\textbf{\textit{cores}}& \\
\hline
%Layers & More table copy$^{\mathrm{a}}$& &  \\
[60,256] & 28 & [100,512] & 15 & [180,1024] & 7 & \\
\cline{1-6} 
[200,64] & 49 & [504,192] & 25 & [968,400] & 10 & 98.64\\
\cline{1-6} 
[240,128] & 18 & [400,256] & 9 & [1008,768] & 3 & \\
\hline
& & & & & & \textbf{Total} \\
& 95 & & 49 & & 20 & \textbf{cores} \\
\hline 
%\multicolumn{4}{l}{$^{\mathrm{a}}$Sample of a Table footnote.}
\end{tabular}
\label{tab3}
\end{center}
\end{table}

\begin{table}[htbp]
\caption{Keeping architecture constant: CIFAR-10 dataset}
\begin{center}
\begin{tabular}{|c|c|c|c|c|c|c|}
\hline
\multicolumn{6}{|c|}{\textbf{Core Size}}&\textbf{Acc} \\
\cline{1-6} 
\multicolumn{2}{|c|}{\textbf{\textit{256$\times$256}}}&\multicolumn{2}{|c|}{\textbf{\textit{512$\times$512}}}&\multicolumn{2}{|c|}{\textbf{\textit{1024$\times$1024}}}&\textbf{(\%)} \\
\cline{1-6}
\textbf{\textit{Core}}&\textbf{\textit{No of}}&\textbf{\textit{Core}}&\textbf{\textit{No of}}&\textbf{\textit{Core}}&\textbf{\textit{No of}}& \\
\textbf{\textit{utilization}}&\textbf{\textit{cores}}&\textbf{\textit{utilization}}&\textbf{\textit{cores}}&\textbf{\textit{utilization}}&\textbf{\textit{cores}}& \\
\hline
%Layers & More table copy$^{\mathrm{a}}$& &  \\
[180,240] & 30 & [300,512] & 14 & [540,1024] & 7 & \\
\cline{1-6} 
[200,64] & 49 & [504,192] & 17 & [968,400] & 10 & 61.41\\
\cline{1-6} 
[240,128] & 18 & [400,256] & 9 & [1008,768] & 3 & \\
\hline
& & & & & & \textbf{Total} \\
& 97 & & 40 & & 20 & \textbf{cores} \\
\hline 
%\multicolumn{4}{l}{$^{\mathrm{a}}$Sample of a Table footnote.}
\end{tabular}
\label{tab4}
\end{center}
\end{table}

\subsubsection{Keeping architecture different}

The different neural network architectures chosen for MNIST and CIFAR-10 datasets for different core sizes are shown in table \ref{tab1} and \ref{tab2}. For this set of experiment, the architecture is changed slightly to fit onto the respective core sizes. The modification of the network is only done on the number of feature maps or channels in different layers. This modification will not really affect the mapping much. But, rather better accuracies are obtained with same number of cores utilized. The convolutional filter size used is 3$\times$3 throughtout the layers. The strides and padding used between all the layers are exactly same as mentioned in the previous subsection. From table \ref{tab5} and \ref{tab6}, the results for MNIST and CIFAR-10 classification accuracy is shown for different core sizes and can be seen that the accuracy improves with increase in core sizes. This is obvious that bigger network can be mapped on to neuromorphic chips with bigger core sizes, bigger the network, better the accuracy. The core utilization varies with mapping but the number of cores utilized remains same with core sizes.

\begin{table}[htbp]
\caption{Keeping architecture different: MNIST dataset}
\begin{center}
\begin{tabular}{|c|c|c|c|}
\hline
\multicolumn{3}{|c|}{\textbf{Core Utilization}}&\textbf{No: of} \\
\cline{1-3}
\textbf{\textit{256$\times$256}}&\textbf{\textit{512$\times$512}}&\textbf{\textit{1024$\times$1024}}&\textbf{(cores)} \\
\hline
[60,256] & [60,512] & [60,1024] & 28 \\
\hline 
[200,64] & [400,128] & [800,256] & 49\\
\hline 
[240,128] & [480,256] & [960,512] & 18 \\
\hline
& & & \textbf{Acc} \\
98.64 & 98.75 & 98.89 & \textbf{\%} \\
\hline 
%\multicolumn{4}{l}{$^{\mathrm{a}}$Sample of a Table footnote.}
\end{tabular}
\label{tab5}
\end{center}
\end{table}

\begin{table}[htbp]
\caption{Keeping architecture different: CIFAR-10 dataset}
\begin{center}
\begin{tabular}{|c|c|c|c|}
\hline
\multicolumn{3}{|c|}{\textbf{Core Utilization}}&\textbf{No: of} \\
\cline{1-3}
\textbf{\textit{256$\times$256}}&\textbf{\textit{512$\times$512}}&\textbf{\textit{1024$\times$1024}}&\textbf{(cores)} \\
\hline
[180,256] & [180,512] & [180,1024] & 30 \\
\hline 
[200,64] & [400,128] & [800,256] & 49\\
\hline 
[240,128] & [480,256] & [960,512] & 18 \\
\hline
& & & \textbf{Acc} \\
61.41 & 64.39 & 66.25 & \textbf{\%} \\
\hline 
%\multicolumn{4}{l}{$^{\mathrm{a}}$Sample of a Table footnote.}
\end{tabular}
\label{tab6}
\end{center}
\end{table}

\section{Discussion and Conclusion}
\label{DaC}

\begin{figure}[htbp]
\centerline{\includegraphics[scale=0.5]{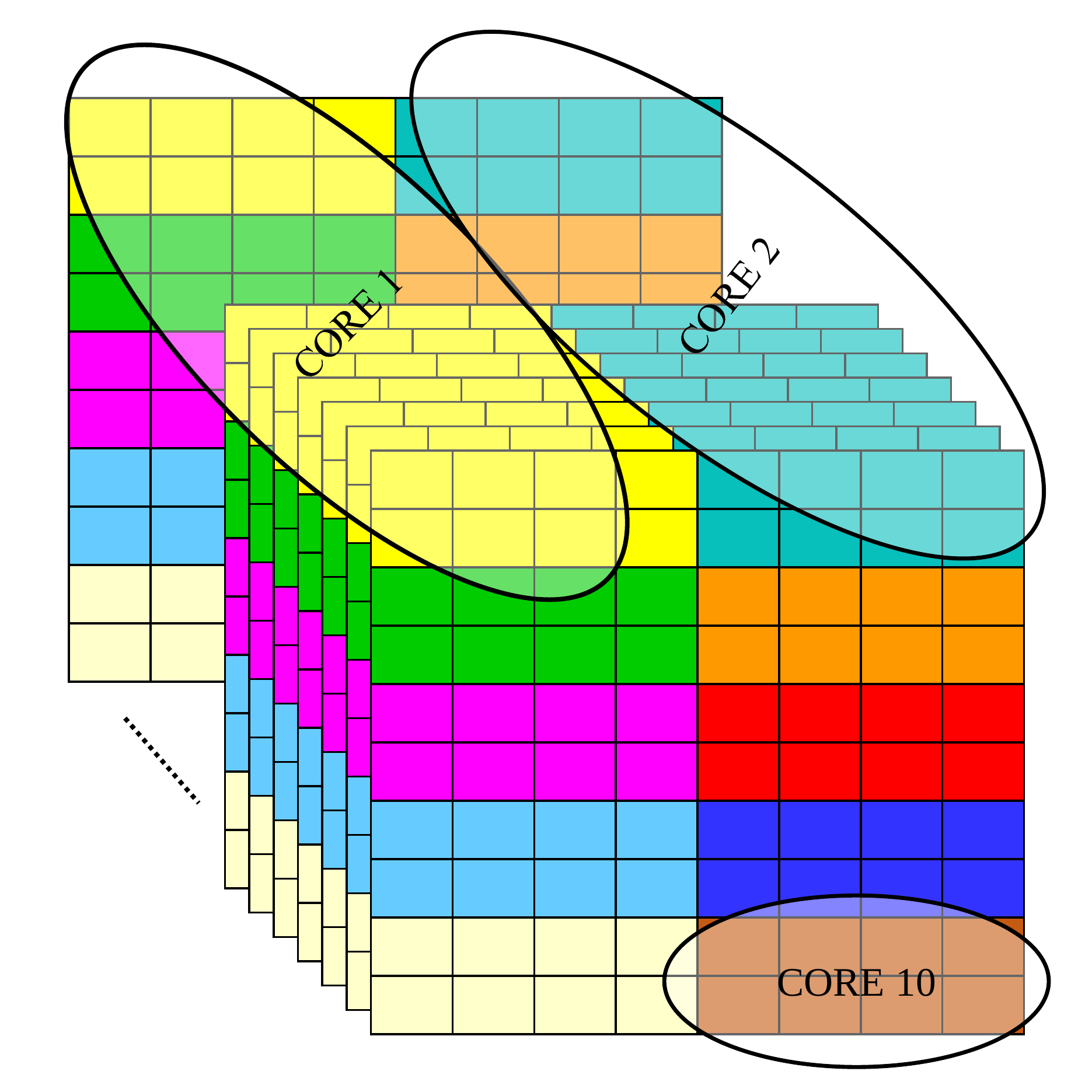}}
\caption{Division of a convolutional neural network layer into different neuromorphic cores.}
\label{Cores_layer}
\end{figure}

Random access memories are popular in terms of in-memory computation. Resistive random access memory (RRAM) became more popular in the field of neuromorphic computing chips with the capability of doing both computation and memory at the same time. These two terminal RRAM devices are very much compatible with the crossbar array of synapses architecture, which enhanced its acceptance in the field of neuromorphic chips. Apart from RRAM, there are other devices like floating-gate MOSFET \cite{Roshan_ijcnn} \cite{Roshan_TNNLS} \cite{Roshan_iscas} \cite{Roshan_t-stdp}, memristors \cite{memristor} \cite{mostafa} \cite{memristor_HRL}, thin-film devices \cite{Rohit} and spin devices \cite{spin} to be the contender of synaptic devices in a neuromorphic chip. %The MaD framework is completely generic in the sense that it can be applied to any crossbar arrays of synapses as in the case of IBM truenorth.

The mapping of different portions of a convolutional layer onto different cores is shown in the fig. \ref{Cores_layer}. Different colors within the layer shows that those neurons are mapped onto particular core. For example, neurons in yellow are mapped onto core 1 and neurons in brown are mapped onto core 10 etc. 

The challenges in mapping onto a single neuromorphic core are mainly explained in the section for optimizations. One of the major priority while mapping is to choose the shape of the neurons in a layer that the chosen neurons and its corresponding axons could map completely onto a neuromorphic core without splitting the matrix between cores. Another concern is to avoid the padded neurons while inferencing or mapping as these padded neurons during training is necessary to keep the size of the input activations but during inference these padded neurons become hardware overhead. In this paper, these two challenges are mitigated using simple techniques in the mapping function.

From the results, it can be seen that bigger the core size, easier to map  a bigger network and better the accuracy. Similarly, if the accuracy is fixed, then the lesser number of cores are utilized in a neuromorphic chip with bigger core size. This is infact better compared to usage of more number of cores in a neuromorphic chip with smaller core sizes because the communication between neuromorphic cores will consume more power than the computations. Eventhough the neuromorphic chip with bigger core size is preferable, the bottleneck is the design possibility of such bigger crossbar array of synapses with the latest CMOS technology. Number of cores in a neuromorphic chip depends on the core size and the available silicon area for the chip. Hence, number of cores and core size become a neuromorphic hardware constraint other than the major hardware constraints like synaptic noise, precision of weight and outputs. This paper gives an overview of mapping in neuromorphic chip with respect to the utilization of number of cores. The python wrapper for MaD framework can output a visual representation of each core in a format easily verifiable by the users (.csv or .xls). The verification of network activations and inferencing becomes quite simple as well. The code for python wrapper can be shared upon request.

\section*{Acknowledgment}

\end{document}